\documentclass[sigconf]{acmart}
\usepackage{float}
\usepackage{multirow}
\usepackage{balance}
\usepackage{hyperref}

\usepackage{xcolor}
\definecolor{crimson}{rgb}{0.86, 0.08, 0.24}
\definecolor{green}{rgb}{0, 0.5, 0.25}
\definecolor{purple}{rgb}{0.75, 0, 1}
\definecolor{orange}{rgb}{1, 0.5, 0.25}
\definecolor{yellow}{rgb}{1, 1, 0}
\definecolor{new_blue}{rgb}{0, 0.5, 1}
\newcommand{\final}[1]{\textcolor{black}{#1}}

\AtBeginDocument{%
  }

\copyrightyear{2023} 
\acmYear{2023} 
\setcopyright{acmlicensed}\acmConference[MM '23]{Proceedings of the 31st
ACM International Conference on Multimedia}{October 29-November 3,
2023}{Ottawa, ON, Canada}
\acmBooktitle{Proceedings of the 31st ACM International Conference on
Multimedia (MM '23), October 29-November 3, 2023, Ottawa, ON, Canada}
\acmPrice{15.00}
\acmDOI{10.1145/3581783.3611789}
\acmISBN{979-8-4007-0108-5/23/10}
\acmSubmissionID{}

\begin{document}

\title{CLIP-Count: Towards Text-Guided Zero-Shot Object Counting}

% \author{ACM MM Submission 412}
% \affiliation{%

% \affiliation{%

% \affiliation{%

\author{Ruixiang Jiang}
\email{rui-x.jiang@connect.polyu.hk}
\affiliation{%
  \institution{The Hong Kong Polytechnic University}
  \city{HKSAR}
  \country{China}
}

\author{Lingbo Liu}
\email{lingbo.liu@polyu.edu.hk}
\affiliation{%
  \institution{The Hong Kong Polytechnic University}
  \city{HKSAR}
  \country{China}}

\author{Changwen Chen}
\authornote{Changwen Chen is the corresponding author. This research is supported by The Hong Kong Polytechnic University (ZVVK-P0036744).}
\email{changwen.chen@polyu.edu.hk}
\affiliation{%
  \institution{The Hong Kong Polytechnic University}
  \city{HKSAR}
  \country{China}}


\renewcommand{\shortauthors}{Ruixiang Jiang, Lingbo Liu, Changwen Chen}

\begin{abstract}
% Recent advances in visual-language models have shown remarkable zero-shot text-image matching ability that is transferable to down-stream tasks such as object detection and segmentation. However, adapting these models for object counting, which involves estimating the number of objects in an image, remains a formidable challenge. In this study, we conduct the first exploration of transferring visual-language models for class-agnostic object counting. Specifically, we propose CLIP-Count, a novel pipeline that estimates density maps for open-vocabulary objects with text guidance in a zero-shot manner, without requiring any finetuning on specific object classes. To align the text embedding with dense image features, we introduce a patch-text contrastive loss that guides the model to learn informative patch-level image representations for dense prediction. Moreover, we design a hierarchical patch-text interaction module that propagates semantic information across different resolution levels of image features. Benefiting from the full exploitation of the rich image-text alignment knowledge of pretrained visual-language models, our method effectively generates high-quality density maps for objects-of-interest. Extensive experiments on FSC-147, CARPK, and ShanghaiTech crowd counting datasets demonstrate that our proposed method achieves state-of-the-art accuracy and generalizability for zero-shot object counting. 

\final{
Recent advances in visual-language models have shown remarkable zero-shot text-image matching ability that is transferable to downstream tasks such as object detection and segmentation. Adapting these models for object counting, however, remains a formidable challenge. In this study, we first investigate transferring vision-language models (VLMs) for class-agnostic object counting. Specifically, we propose \textbf{CLIP-Count}, the first end-to-end pipeline that estimates density maps for open-vocabulary objects with text guidance in a zero-shot manner. To align the text embedding with dense visual features, we introduce a patch-text contrastive loss that guides the model to learn informative patch-level visual representations for dense prediction. Moreover, we design a hierarchical patch-text interaction module to propagate semantic information across different resolution levels of visual features. Benefiting from the full exploitation of the rich image-text alignment knowledge of pretrained VLMs, our method effectively generates high-quality density maps for objects-of-interest. Extensive experiments on FSC-147, CARPK, and ShanghaiTech crowd counting datasets demonstrate state-of-the-art accuracy and generalizability of the proposed method. Code is available: \href{https://github.com/songrise/CLIP-Count}{https://github.com/songrise/CLIP-Count.}}
\end{abstract}




\maketitle
\begin{figure}[t]
    \centering
    \includegraphics[width=\linewidth]{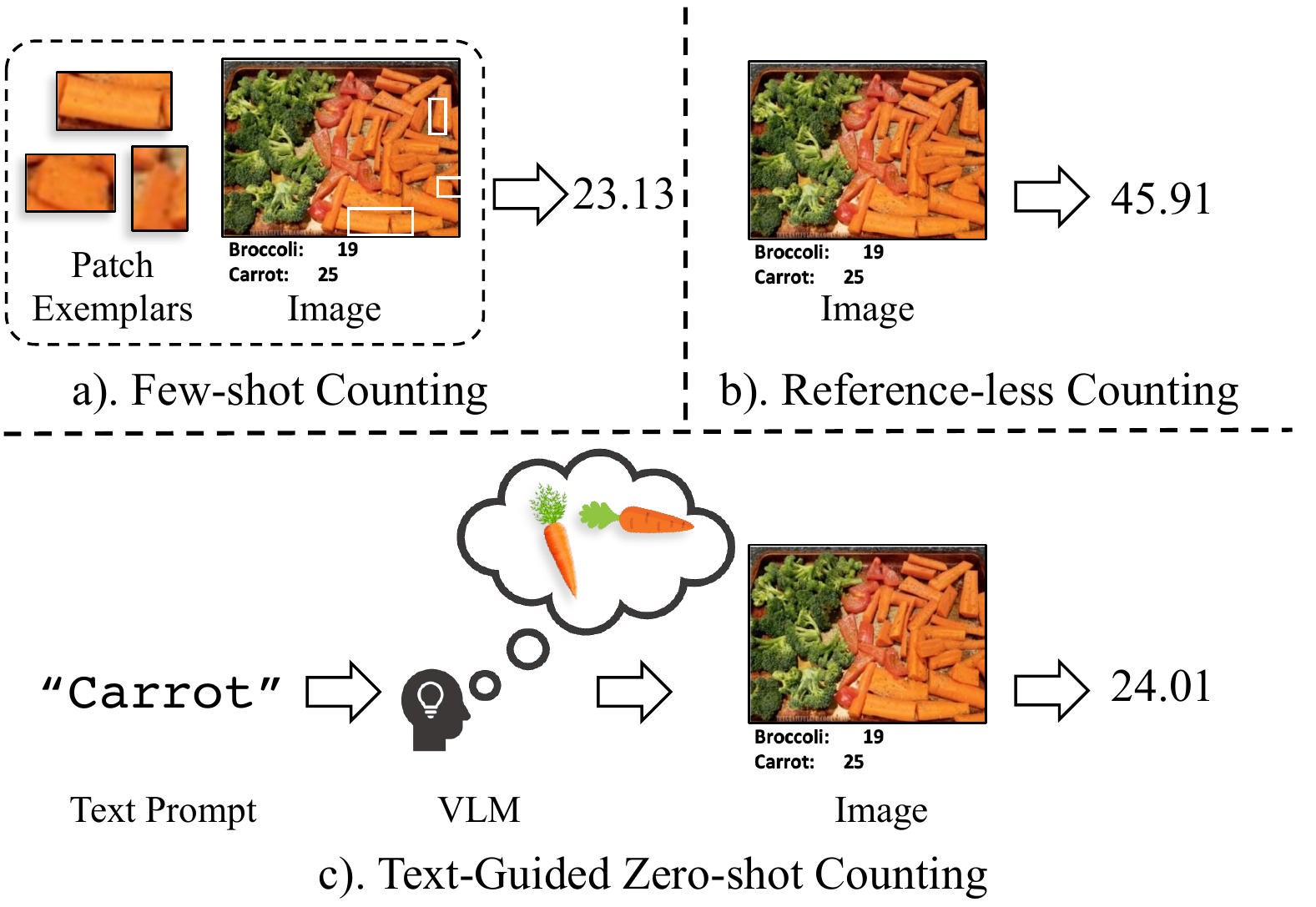}
    % \caption{Illustration of different schemes of class-agnostic object counting. Few-shot counting requires manually-labeled image patches to specify the objects of interest, while reference-less counting mines and counts salient objects automatically, but it may fail to recognize objects of interest. In this work, we focus on the more flexible text-guided zero-shot counting, which uses text prompts to recognize and count objects of interest via pretrained Vision-Language Model (VLM), without relying on extra manual annotations.
    % } 
     \caption{\final{Illustration of different schemes of generalized object counting. a) Few-shot counting requires manually-labeled image patches for training and inference. b) Reference-less counting mines and counts salient objects automatically, but it cannot specify objects-of-interest. c) Zero-shot counting uses text prompts to specify and count objects of interest via pretrained VLM, without relying on manual annotations.}}
    % } 
    \label{fig:diff}
\end{figure}

\section{Introduction}
Over the past decade, object-specific (\textit{e.g.}, crowd and vehicle) counting has been extensively studied \cite{sindagi2017generating,zhang2017fcn,liu2018crowd,qiu2019crowd,liu2019crowd,yuan2020crowd,liu2020efficient,liu2021cross,li2022video,han2022dr,wu2022multimodal}. Despite the progress, it requires training a specific network with massively labeled samples for each class of objects. In contrast, human counting ability is agnostic to specific object classes. When presented with previously unseen objects, we identify, compare, and treat similar objects as the counting target \cite{dehaene2011number}. This observation has motivated the \final{emergence} of class-agnostic object counting algorithms \cite{lin2021object,xu2023zero,zhou2022learning,djukic2022low,liu2022countr, lu2018class,ranjan2022exemplar,hobley2022learning,wang2023gcnet}, which aim at training a unified/shared model to estimate the number of arbitrary objects-of-interest in the given image, as shown in Fig.~\ref{fig:diff}-(a). By annotating a few image patches as exemplars and computing the similarities between exemplars and image regions, these methods have achieved good generalizability and counting accuracy. 

However, most previous class-agnostic counting methods assume that accurate bounding boxes of exemplars are easily available both during training and inference, which is not always the case in real-world applications. Therefore, in practice, they usually request users to manually annotate the exemplars of objects-of-interest, which is unfriendly to end-users. Moreover, even when exemplars are annotated, the high intra-class variance of query objects can still lead to biased object counts \cite{liu2022countr,xu2023zero}. To address these issues, reference-less \final{counting} methods have been proposed to automatically mine and count salient objects at inference time  \cite{ranjan2022exemplar,hobley2022learning}. While these approaches avoid cumbersome manual annotation, they lack the discriminative ability to specify the object class of interest in the presence of multiple object classes, as illustrated in  Fig.~\ref{fig:diff}-(b). Overall, the limitations of existing counting schemes highlight the need for more flexible and robust guidance in generalized object counting.

In this work, we challenge the conventional notion that self-similarity between exemplars and \final{the query} image is essential for generalized object counting. To this end, we propose a new counting scheme called text-guided zero-shot object counting, which utilizes an additional natural language prompt as input to query object count in given images, as shown in Fig.~\ref{fig:diff}-(c). In particular, the use of text prompts as guidance offers two main benefits. Firstly, it reduces the need for manual annotation of patch exemplars both during training and testing, making it more user-friendly and scalable to larger datasets. Secondly, text prompts offer greater flexibility compared to patch exemplars, as they can cover both general descriptions such as \textit{"food"} and specific descriptions such as \textit{"red apple in a basket"}. Therefore, our text-guided scheme is more flexible and promising for class-agnostic object counting.

However, using natural language to guide generalized object counting presents several challenges. Firstly, unlike patch annotations that provide explicit appearance and shape information, text prompts only contain implicit descriptions of the query object, which can be intrinsically ambiguous. 
Additionally, text prompts are expected to precisely locate objects-of-interest in input images, and the effective semantic alignment of these two modalities is non-trivial.   
Moreover, achieving zero-shot object counting requires the model to be generalizable to open-vocabulary text prompts and object classes, but the lack of large-scale annotated datasets severely hinders the development and evaluation of text-guided zero-shot object counting methods.

Taking the above issues into consideration, we propose a unified text-guided zero-shot object counting method, termed \textbf{CLIP-Count}. Specifically, our method is developed based on Contrastive Language-Image Pre-training \cite{radford2021learning} (CLIP), which endows our model with zero-shot image-text alignment ability. To transfer the powerful image-level CLIP to dense tasks such as density estimation, we first design a patch-text contrastive loss to align text and patch embedding space. This alignment is achieved by simultaneously tuning visual~\cite{jia2022visual} and text prompts~\cite{zhou2022learning,zhou2022conditional} to enable parameter-and-data-efficient transfer learning of pretrained CLIP. To propagate textual information to dense image features for text-guided counting, we further elaborate a hierarchical patch-text interaction module that correlates text and image to different resolutions. Thanks to those tailor-designed modules, our method can effectively exploit the rich image-text alignment knowledge of CLIP to generate high-quality density maps for objects-of-interest. Extensive experiments on FSC-147~\cite{ranjan2021learning}, CARPK~\cite{hsieh2017drone}, and ShanghaiTech crowd counting~\cite{zhang2016single} dataset demonstrate that our method is effective and generalizable to unseen object classes and text prompts.

In summary, our major contributions are as follows:
\begin{itemize}
    \item To the best of our knowledge, our CLIP-Count is the first end-to-end
    text-guided zero-shot object counting model, which is flexible and scalable in real-world applications.  
    \item  We propose a novel approach whereby visual prompts and text prompts are learned simultaneously to fully exploit the rich pre-trained knowledge of CLIP.
    \item \final{we propose two tailored modules (\textit{i.e.}, patch-text contrastive loss and hierarchical patch-text interaction module) to improve the counting ability of CLIP.}
    \item Extensive experiments conducted on three object counting datasets demonstrate the state-of-the-art \final{accuracy and generalizability} of the proposed CLIP-Count method.
\end{itemize}

\begin{figure*}
    \centering
    \includegraphics[width=0.8\linewidth]{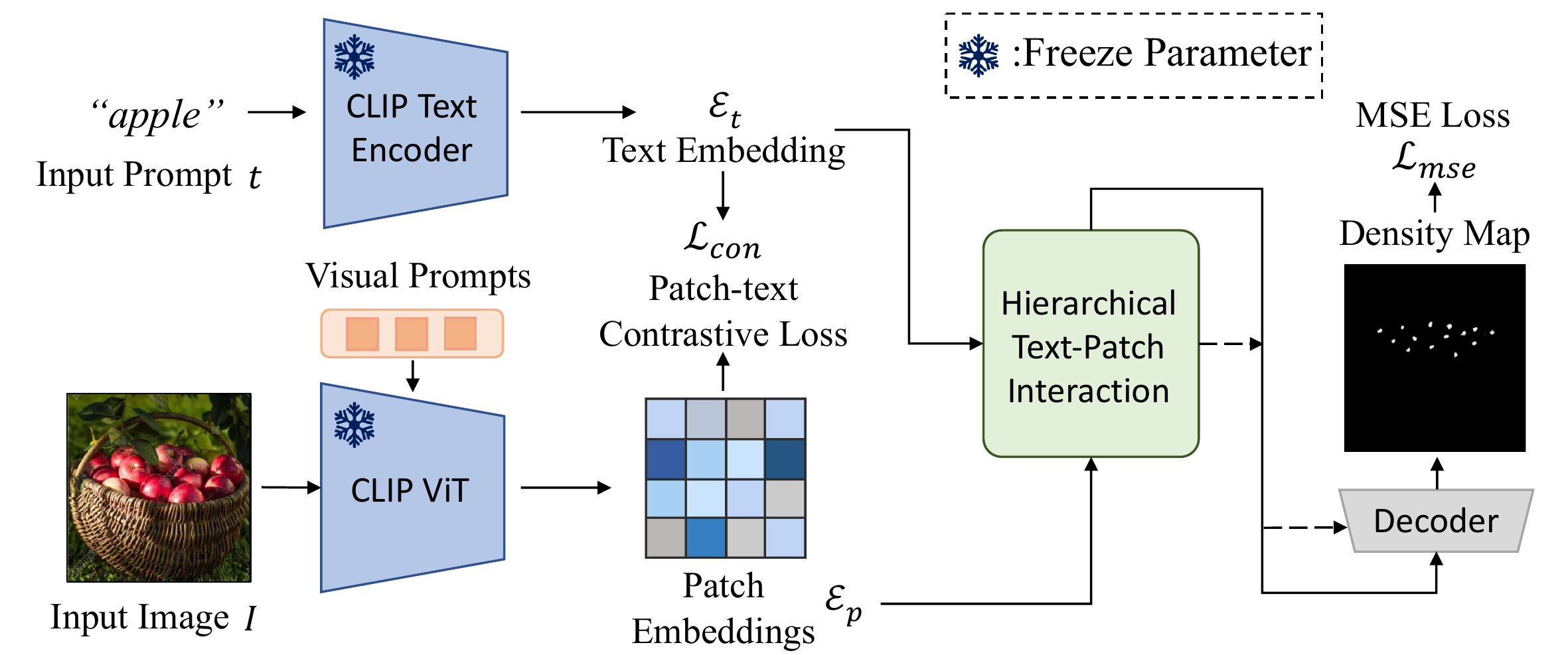}
    \vspace{-3mm}
    \caption[Overview of our CLIP-Count.]{\textbf{Overview of our CLIP-Count.} We freeze both encoders in pretrained CLIP and append a small amount of learnable prompts to transfer the pretrained knowledge. A patch-text contrastive loss is employed to align dense visual features with text. To interact the two modalities, we design a hierarchical text-patch interaction module that outputs \final{two} multi-modal feature maps at \final{different} resolutions. Finally, we decode the multi-modal feature maps with a CNN decoder. }
    \label{fig:over}
\end{figure*}

\section{Related Works}
\subsection{Few-shot Object Counting}
Few-shot object counting algorithms aim to learn a generalized model that counts anything with exemplar as inference-time guidance. The pioneering work, GMN \cite{lu2018class} first formulates class-agnostic counting as a matching problem to exploit the self-similarity in counting tasks. Following this work, FamNet \cite{ranjan2021learning} learns to predict density maps with ROI pooling, and they further introduce a new dataset for class-agnostic counting, namely FSC-147 \cite{ranjan2021learning}. \final{Subsequent advancements can be categorized into two streams}. One approach \final{involves utilizing} more advanced visual backbones, such as vision transformers \final{(ViT)}, to enhance the extracted feature representation (\textit{e.g.},\ CounTR~\cite{liu2022countr}, LOCA~\cite{djukic2022low}). The second idea is to enhance the exemplar matching process by explicitly modeling exemplar-image similarity (\textit{e.g.},\ BMNet~\cite{shi2022represent}, SAFECount~\cite{you2023few}) or by further exploiting exemplar guidance~\cite{djukic2022low, lin2022scale,wang2023gcnet}. Despite their good performance, all of those methods require additional patch-level annotation at training and inference time, which could be costly to obtain.

\subsection{Reference-less and Zero-shot Object Counting} 
Reference-less counting has recently emerged as a more promising direction for class-agnostic counting without human annotation. The earliest attempt, RepRPN-Counter \cite{ranjan2022exemplar}, proposes a region proposal module to extract the salient object in replacement of exemplar input. RCC \cite{hobley2022learning} leverages pretrained ViT \cite{dosovitskiy2020image, caron2021emerging} to implicitly extract the salient objects, and directly regress a scalar as the estimated object count. Notably, some recent few-shot counting models \cite{liu2022countr, djukic2022low,wang2023gcnet} could also be configured to perform reference-less counting. While those methods do not need exemplars, they fail to provide a way for specifying object-of-interest in presence of multiple object classes. Concurrent with this work, Xu \textit{et al.} ~\cite{xu2023zero} introduces the task of zero-shot object counting, where only the class name is needed in inference time. They employ a two-stage training scheme: in the first stage, they train a few-shot object counter with exemplar supervision, and to achieve zero-shot counting, they additionally train a text-conditional variational autoencoder (VAE) on a closed-set of objects to generate exemplar prototypes. While their work shares similarities with \final{ours} at a high level, the key difference is that their two-stage training scheme still necessitates patch exemplars \final{for training}. In contrast, our method \final{could be trained end-to-end without patch-level supervision}.

\subsection{Vision-Language Pre-training} 
VLMs pre-trained on extremely large dataset demonstrate advanced zero-shot image-text matching ability~\cite{radford2021learning,li2022blip,jia2021scaling}. In particular, CLIP~\cite{radford2021learning} learns an aligned multi-modal embedding space, and it has inspired various applications that use it as an image-level classifier ~\cite{hong2022avatarclip,wang2022nerf,zhang2022pointclip,patashnik2021styleclip}. To further exploit the powerful image-level representation in CLIP to perform dense tasks, DenseCLIP~\cite{rao2022denseclip} elaborates a context-aware prompting technique to facilitate dense prediction such as segmentation. Concurrently, RegionCLIP~\cite{zhong2022regionclip} identifies the lack of localization ability issue in CLIP, and they distill pre-trained knowledge in CLIP to enable fine-grained image-region matching. More recently, ZegCLIP \cite{zhou2022zegclip} designs a one-stage framework that calculate object masks based on patch embeddings for segmentation, while GridCLIP~\cite{lin2023gridclip} aligns grid-level representation with text to enable one-stage object detection. The majority of those improvements focus on segmentation and detection. Very recently, two CLIP-based object counting models were proposed~\cite{paiss2023teaching,liang2023crowdclip}, both based on image-level classification and limited in counting granularity and accuracy. Currently, it remains a challenge to achieve \final{density estimation with VLMs.}
\section{Method}

In this section, we present our proposed method, CLIP-Count, for text-guided zero-shot object counting.  We will first review the structure of CLIP in Sec.~\ref{sec: baseline prelim}. The objective of zero-shot class-agnostic object counting is defined in Sec.~\ref{sec:objective}. We elaborate on the design of our proposed method in Sec.~\ref{sec:contra}, and Sec.~\ref{sec:adapt}.

\subsection{Preliminary: CLIP}\label{sec: baseline prelim}
% In this section, we briefly review the structure of CLIP~\cite{radford2021learning}, which serves as the backbone of CLIP-Count.

CLIP~\cite{radford2021learning} connects image and language through large scale pre-training on image and text pairs. It learns to align the text and image representations, so that the similarity of image and text pair could be calculated by dot product of their respective embeddings. Formally, let $\mathcal{E}_{I} \in \mathbb{R}_n$ and $\mathcal{E}_{t}\in \mathbb{R}_n$ denote the embedding of an image $I$ and a text $t$, where $n=512$ is the dimension of CLIP embedding space. The cosine similarity of the text and image pair could be calculated as:
\begin{equation}
\operatorname{CLIP}(I,t) = \frac{\mathcal{E}_{t}\cdot \mathcal{E}_{I}}{||\mathcal{E}_{t}||||\mathcal{E}_{I}||}
\label{eq: clip}
\end{equation}

CLIP employs two encoders for encoding the image and text. Specifically, the last layer of ViT-based CLIP visual encoder~\cite{dosovitskiy2020image} outputs a global image feature $z_{0}\in \mathbb{R}_{d}$ and a 1/16 resolution patch-level feature map $\mathbf{z}_{x}\in \mathbb{R}_{p^2\times d}$, where $p=14$ denotes the number of image patches per column and row, and $d=768$ represents ViT latent dimension. CLIP applies a linear projection $\phi_z : \mathbb{R}_d \mapsto \mathbb{R}_n $ over the global feature to get the image embedding $\mathcal{E}_{I} = \phi_z(z_0)$, while the patch-level feature map $\mathbf{z}_x$ is discarded. In this work, we focus on utilizing $\mathbf{z}_x$ for density estimation.

\subsection{Objective}\label{sec:objective}
 Our objective is to count anything with text guidance. Formally, given an image $I\in \mathbb{R}_{H\times W\times 3}$ that contains arbitrary object class(es) as well as a natural language prompt $t$ that specifies the object(s) of interest, our objective is to estimate a density map $\hat y \in \mathbb{R}_{H\times W} $ for the specified object(s). The estimated object count could be calculated by summing the density map $N_{pred} = \operatorname{SUM}(\hat y)$.

 We denote the object classes contained in each set as $\mathcal{C}_{train},\mathcal{C}_{val},\\ \mathcal{C}_{test}$. Under the class-agnostic setting, the testing and validation sets do not overlap with the training set: $\mathcal{C}_{train} \cap \mathcal{C}_{val} = \varnothing$ and $\mathcal{C}_{train} \cap \mathcal{C}_{test} = \varnothing$. We optimize the model $F_{\theta}(I,t)=\hat{y}$ on the training set using ground truth density map supervision $y$, and evaluate its performance on the validation and testing sets.

It is worth noting that incorporating text guidance at inference time is a standard practice both for zero-shot object counting~\cite{xu2023zero} and for other text-driven tasks with CLIP~\cite{zhou2022zegclip,hong2022avatarclip, sanghi2022clip,wang2022nerf}. Additionally, previous research~\cite{liu2022countr,djukic2022low,hobley2022learning} use the terms ``reference-less counting'' and ``zero-shot counting'' interchangeably, despite their inherent differences. For clarity, we refer to the methods that solely utilize image input as "reference-less", while those that incorporate text as 
 guidance are referred to as "zero-shot counting". Fig.~\ref{fig:diff} summarizes their difference.

\subsection{Aligning Text with Dense Visual Features}\label{sec:contra}
 
 \begin{figure}
    \centering
    \includegraphics[width=\linewidth]{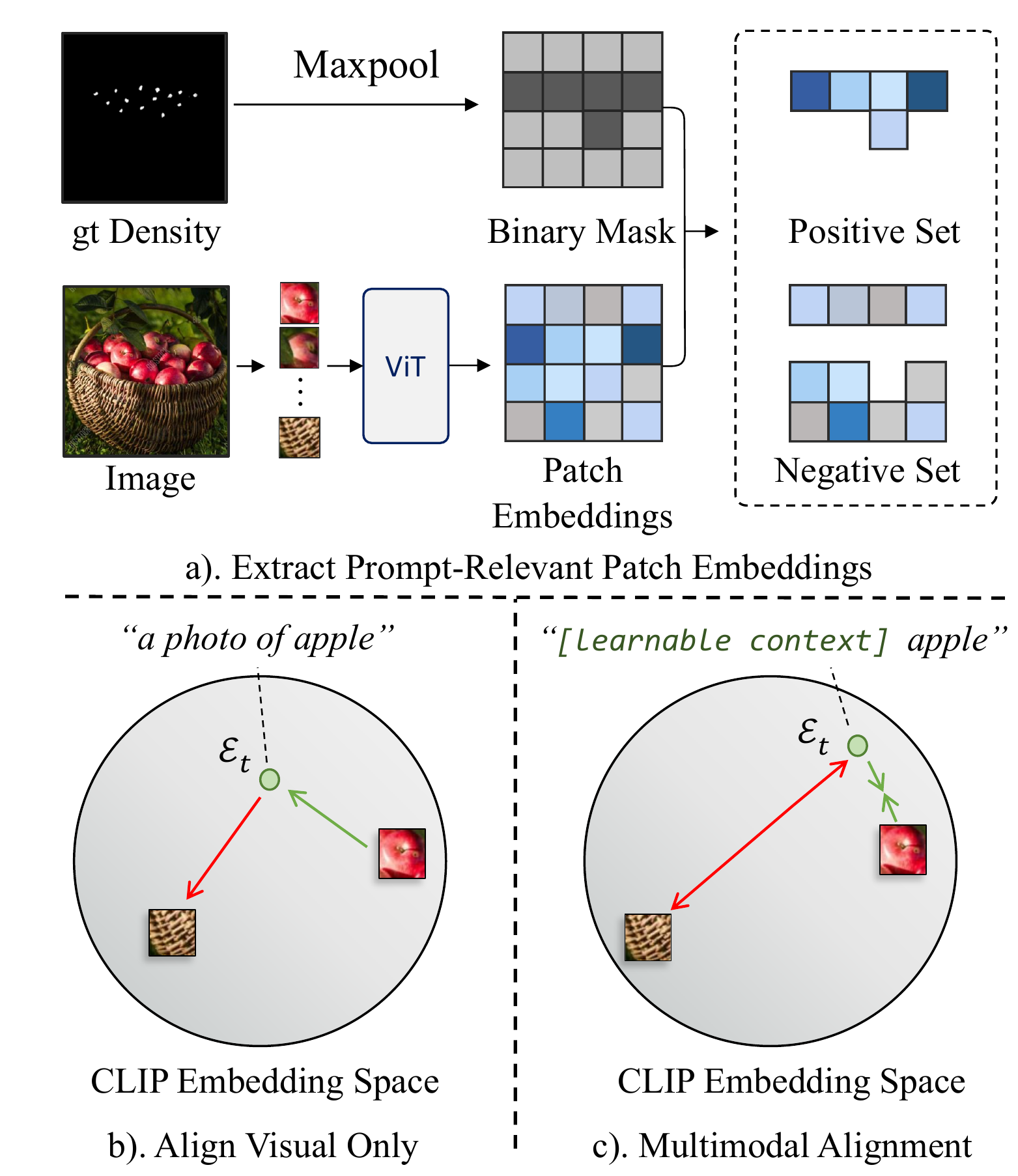}
    \caption{Illustration of patch-text contrastive loss. a.) we utilize g.t. density map to determine the objectness at patch-level, and we split the patch embeddings into two sets accordingly; b) finetune the visual encoder to align patch embeddings to the fixed text embedding space; c) shift both text and patch embeddings. The number of patches in a) and the two image patches in b) and c) are only for illustration purposes.}
    \label{fig:contrast}
\end{figure}

CLIP has demonstrated remarkable performance in measuring global image-text similarity. However, transferring such ability to pixel level for dense-prediction is non-trivial, as the vanilla CLIP intrinsically lacks localization ability~\cite{rao2022denseclip,li2023clip}, which limits its performance in object detection, segmentation, and counting. To alleviate this limitation, we propose adapting the CLIP vision encoder to enhance the localization ability of patch-level feature map $\mathbf{z}_x$.

To be more specific, we propose a patch-text contrastive loss that maximizes the mutual information between text prompt and prompt-relevant patch-level features. To achieve this, we first apply a linear projection \final{$\phi_p : \mathbb{R}_d \mapsto \mathbb{R}_n$} to map \final{$\mathbf{z}_x$} to the same \final{channel} dimension as the text embedding $\mathcal{E}_t$:
 \begin{equation}
     \{\mathcal{E}_p^i =\phi_p(\mathbf{z}_x^i)| i = 1,2,\cdots, p^2\}
 \end{equation}
 
For brevity, we call the reshaped $\mathbf{\mathcal{E}}_p\in \mathbb{R}_{p\times p\times n}$ as \textit{patch embeddings} of the input image. It is worth noting that this naming differs from the convention used in ViT, as our $\mathcal{E}_p$ encodes cross-patch information due to the image-level perceptive field in global attention~\cite{dosovitskiy2020image}.  To determine the location of objects, we maxpool the ground truth density map $y$ to get a patch-level binary mask \final{$\mathcal{M}\in \mathbb{R}_{p\times p}$} for objectness. We then define $\mathcal{P}$ and $\mathcal{N}$ as the positive and negative sets of patch embeddings, depending on their corresponding mask value. Similar as ~\cite{wang2022cris,shi2022represent}, we introduce the following InfoNCE-based~\cite{oord2018representation} contrastive loss:
 \begin{equation}
    \mathcal{L}_{con}=    -\log \frac{\sum_{i\in\mathcal{P}}\exp(s(\mathcal{E}_p^i,\mathcal{E}_t)/\tau)}{\sum_{i\in\mathcal{P}}\exp(s(\mathcal{E}_p^i,\mathcal{E}_t)/\tau)+\sum_{k\in\mathcal{N}}\exp(s(\mathcal{E}_p^k,\mathcal{E}_t)/\tau)}
    \label{eq: contra}
 \end{equation}
where $s(.,.)$ denotes cosine similarity, and $\tau=0.07$ is the temperature parameter~\cite{oord2018representation}.  The contrastive loss function in Eqn.~\ref{eq: contra} aligns the patch embedding space with the text embedding space by pulling positive patch embeddings closer to the fixed text embedding and pushing negative patches further away, \final{which is analogous} to the pre-training objective of CLIP~\cite{radford2021learning}. \final{Furthermore}, we find it beneficial to adjust both the patch and text embeddings simultaneously, as the prompt templates used by CLIP such as \textit{``a photo of \{\textbf{class}\}''} typically describe the image as a whole, rather than at the patch level. Inspired by CoOp~\cite{zhou2022learning}, we optimize continuous token embeddings as the context of input prompt in an end-to-end manner. The entire procedure is summarized in Fig.~\ref{fig:contrast}.

 \begin{figure}
    \centering
    \includegraphics[width=0.8\linewidth]{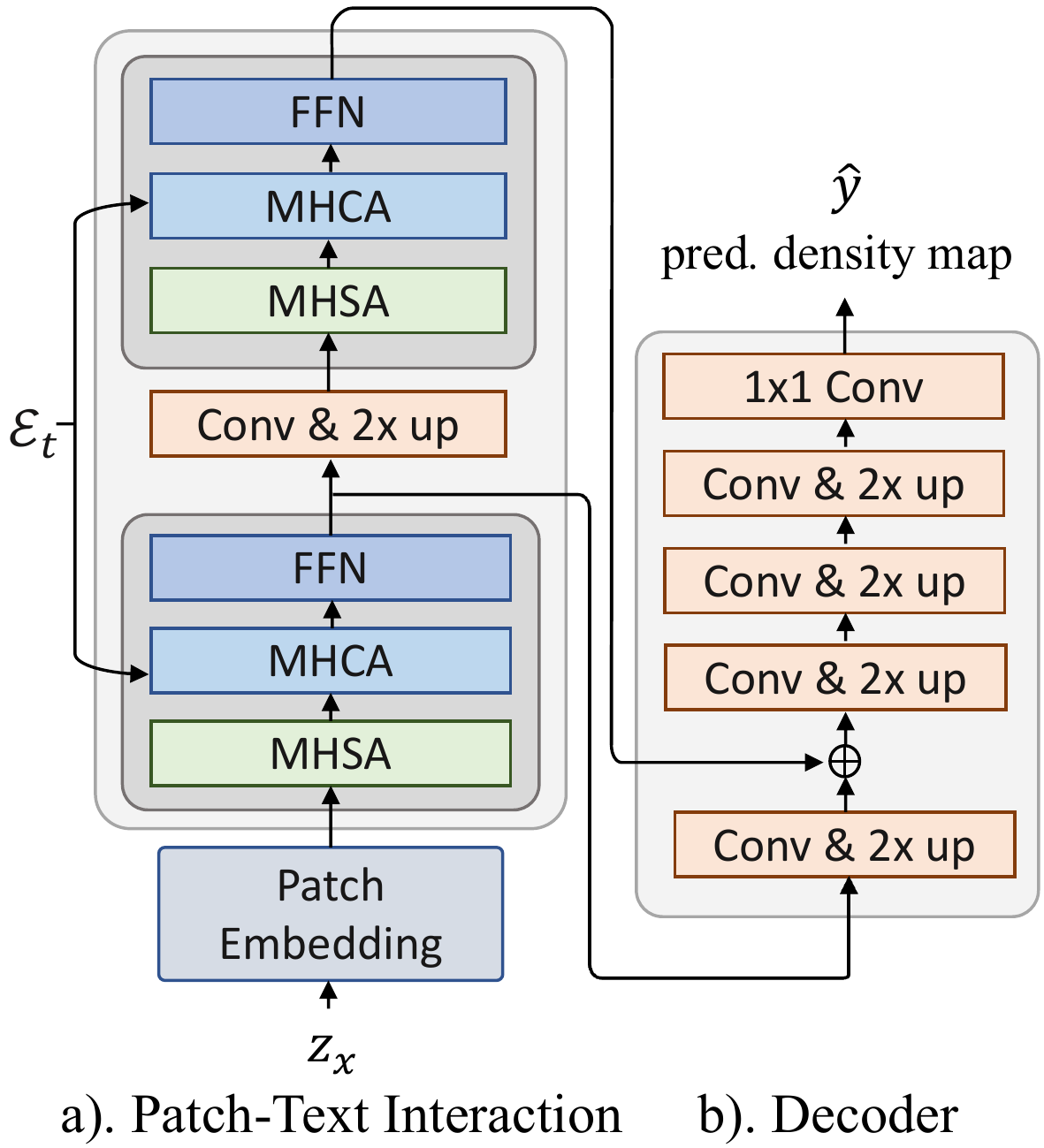}
    \caption{Architecture of the proposed hierarchical patch-text interaction module and the decoder. For brevity, we omit layer norms, positional encoding, and activation functions.}
    \label{fig:decoder}
\end{figure}

\subsection{Adapting CLIP for Density Estimation}\label{sec:adapt}
Our patch-text contrastive loss aims to enhance the patch-level representation and improve the localization ability of CLIP. Nonetheless, two key questions remain: 1) How should the visual encoder be fine-tuned? and 2) How can the patch-level feature map be decoded into a text-conditional density map? In this section, we will address these questions.

\textbf{Visual Prompt Tuning.}  Our model involves transferring the pretrained knowledge in CLIP visual transformer for dense prediction. However, the conventional \textit{``pre-training + finetuning''} paradigm would be computationally expensive as it requires updating the entire model. Furthermore, finetuning a transformer usually requires a significant amount of data, which may not be practical when annotations are expensive to obtain, as is often the case with object counting. To effectively exploit the pretrained knowledge in CLIP, we instead freeze the parameters in CLIP ViT, and concatenate a small amount of trainable parameters as visual prompts~\cite{jia2022visual} to the input of each transformer layer. This allows us to transfer the knowledge in CLIP using a much smaller amount of data and memory while still benefiting from the rich visual representation learned in CLIP. We refer readers to the original paper on visual prompt tuning (VPT)~\cite{jia2022visual} for further details.

\textbf{Hierarchical Text-Patch Interaction Module.} \final{Objects in a scene} usually span a variety of scales, yet the \final{homogenous} transformer structure intrinsically lacks the inductive bias for modeling scale-invariant features~\cite{li2022exploring}, resulting in inaccurate density estimation. To address this issue, we propose a lightweight hierarchical transformer with cross-attention to enable the propagation of text information to different scales of the image feature.  Our interaction module comprises two similar \final{blocks}, where each \final{block} sequentially applies multihead self-attention (MHSA), multihead cross-attention (MHCA), and a two-layer MLP (FFN). The MHSA captures long-distance relationships, while the MHCA propagates semantic information embedded in text to visual features. In between the two layers, a convolution layer with a skip connection is employed, followed by a $2\times$ bi-linear interpolation layer that doubles the resolution of the intermediate multi-modal feature map $M_c\in \mathbb{R}_{p\times p\times n}$. The motivation is to capture \final{the} relationship between text and image at a finer granularity~\cite{li2022exploring}. The final output of the interaction module are the coarse $M_c$ and fine $M_f \in \mathbb{R}_{2p\times 2p\times n}$ multi-modal feature maps. \final{As validated in the Appendix, we find the two-scale design provides the best balance between cost and accuracy.} The entire \final{interaction} process is visualized in Fig.~\ref{fig:decoder}-(a).

\textbf{Density Map Regression.} We decode the multi-modal feature maps $\{M_c,M_f\}$ using a CNN-based decoder, as illustrated in Fig.~\ref{fig:decoder}-(b). The decoder consists of several convolutional \final{layers} and $2\times$ interpolation layers that sequentially \final{doubles} the spatial resolution and decrease the \final{channel} dimension \final{by half}. To decode the two-scale feature maps from the interaction module, we first perform convolution on the coarse multi-modal map $M_c$, and fuse it with $M_f$ before the second convolution by summation:
\begin{equation}
\label{eq: hierachy}
M_f^\prime = 
    \operatorname{Lerp}(\sigma(\operatorname{Conv}_{3\times3}(M_c))) + 
    \sigma(\operatorname{Conv}_{1\times1}(M_f))
\end{equation}
where $\operatorname{Lerp}(.)$ denotes bi-linear interpolation, $\sigma(.)$ is the activation function. 
The final output of the decoder \final{$\hat{y}$} is obtained through a $1\times1$ convolution layer with sigmoid activation.

\section{Experiments}\label{sec: baseline exp}

\subsection{Dataset}

\textbf{FSC-147.} FSC-147~\cite{ranjan2021learning} is a dataset recently proposed for class-agnostic object counting. It consists of 6,135 images across 147 object classes, with non-overlapping object classes in each set.  For each training image, the dataset provides its class name, dot supervision, and three random bounding boxes of objects (\textit{i.e.}, the exemplar patch annotation).  In our experiments, we use the class name as the text prompt $t$, and \final{we} do not use the patch annotation.

\textbf{CARPK.} CARPK~\cite{hsieh2017drone} provides 1448 birds-eye view image of parking lots, containing a total of 89,777 cars. We use CARPK to evaluate the cross-dataset transferability of \final{CLIP-Count}.

\textbf{ShanghaiTech.} The ShanghaiTech crowd counting dataset~\cite{zhang2016single} is a comprehensive dataset for crowd counting. It comprises two parts, A and B, with a total of 1,198 annotated images. Part A consists of 482 images, with 400 designated for training and 182 for testing. Part B comprises 716 images, with 400 for training and 316 for testing. Notably, the two parts were collected using different methods, which presents a challenge for cross-part evaluation.

\subsection{Implementation Details} \textbf{Architecture Detail.} We use OpenAI pre-trained CLIP with ViT-B/16 backbone. For the visual encoder, we use the deep version of VPT, with $20$ visual prompts in each layer. For context learning, we use $2$ prefix tokens. \final{The maxpooling in contrastive loss adopts a kernel size of $16\times 16$ and stride $= 16$, no padding is used}.  For \final{the} density decoder, we use $3\times3$ convolution kernel with stride $=1$ and padding $=1$ to keep spatial resolution. GeLU non-linearity is used after each convolution layers.

\textbf{Training Detail.} The model is trained on the training set of FSC-147. We view our proposed contrastive loss as a pretext task to improve feature representation. Therefore, we \final{first} pre-train the model with contrastive loss \final {only} for 30 epochs before further train it with mean square error (MSE) loss for 200 epochs. The MSE loss is defined as \final{
$
\mathcal{L}_{MSE} =  ||y-\hat{y}||^2 /{(H\times W)}$.}

During both training stages, we use a batch size of 32, \final{and use AdamW \cite{loshchilov2017decoupled} optimizer} with a learning rate of $1\times 10^{-4}$, which decays by a factor of $0.33$ after 100 epochs. The whole training process takes approximately 3 hours on a single Nvidia RTX-3090Ti GPU. We apply the same data augmentations as used in CounTR \cite{liu2022countr}, except for their proposed mosaic augmentation. Additionally, to comply with the input assumption of CLIP, we downsample the training images to $224\times224$, whereas previous generalized counting methods on FSC-147 typically use a resolution of $384\times384$.

% \textbf{Inference.} We test the model on the validation and test set of FSC-147. When inference, we use sliding windows with stride = 128, and the number of object for overlapping area are taken as average.

\subsection{Evaluation Metrics}
Following previous class agnostic counting methods~\cite{liu2022countr,ranjan2021learning,hobley2022learning,djukic2022low,xu2023zero}, we evaluate the performance by mean absolute error (MAE) and root mean squared error (RMSE).
\begin{equation}
    \operatorname{MAE} = \frac{1}{N_I}\sum_{i=1}^{N_I}|N_{pred}^i - N_{gt}^i|,  \
    \operatorname{RMSE} = \sqrt{\frac{1}{N_I}\sum_{i=1}^{N_I}(N_{pred}^i - N_{gt}^i)^2
}
\end{equation}
where $N_I$ denotes the number of images in testing set, and $N_{pred}$, $N_{gt}$ is the predicted and ground truth object count.

\begin{table*}[t]
\centering
\caption{Comparison with state-of-the-art methods on FSC-147. (*) denotes adopting modification for reference-less counting as described in RCC \cite{hobley2022learning}. We highlight the best result for each scheme in bold.}
\label{tab:quantres}
\resizebox{1.6\columnwidth}{!}{%
\begin{tabular}{cccccccc}
\toprule
%Method                       & Source                        & \#shot & \multicolumn{2}{c}{\begin{tabular}[c]{@{}c@{}}val set\\ MAE RMSE\end{tabular}} & \multicolumn{2}{c}{\begin{tabular}[c]{@{}c@{}}test set\\ MAE RMSE\end{tabular}} \\ 

\multirow{2}{*}{Scheme} & \multirow{2}{*}{Method} & \multirow{2}{*}{Source} & \multirow{2}{*}{\#Shot} & \multicolumn{2}{c}{Val Set} & \multicolumn{2}{c}{Test Set} \\
\cline{5-8}
& & & & MAE & RMSE & MAE & RMSE \\
\hline

\multirow{4}{*}{Few-shot} & FamNet~\cite{ranjan2021learning}                       & CVPR2021                      & 3      & 24.32                                  & 70.94                                 & 22.56                                  & {101.54}             \\

& CFOCNet~\cite{yang2021class}  & WACV2021 & 3  &  21.19 & 61.41 & 22.10 & 112.71 \\
& CounTR~\cite{liu2022countr}                       & BMVC2022                      & 3      & 13.13                                  & 49.83                                 & 11.95                                  & \multicolumn{1}{c}{91.23}              \\
& LOCA~\cite{djukic2022low}                         & arXiv2022                     & 3      & \textbf{10.24}                                 & \textbf{32.56}                                 & \textbf{10.97}                                 & {\textbf{56.97}}              \\
& FamNet~\cite{ranjan2021learning}                        & CVPR2021                      & 1      & 26.05                                  & 77.01                                 & 26.76                                  & {110.95}             \\ \hline
%\multicolumn{7}{c}{reference-less}                                                                                                                                                                                                            \\ \hline
\multirow{5}{*}{Reference-less} & FamNet$^*$~\cite{ranjan2021learning}                        & CVPR2021                      & 0      & 32.15                                  & 98.75                                 & 32.27              & 131.46                                 \\ 
 & RepRPN-C~\cite{ranjan2022exemplar} & ACCV2022 & 0      &  29.24            & 98.11             & 26.66              & 129.11                                \\
 & CounTR~\cite{liu2022countr}                       & BMVC2022                      & 0      & 18.07                                  & 71.84                                 & {\textbf{14.71}}              & 106.87                              \\
 & LOCA~\cite{djukic2022low}                         & arXiv2022                     & 0      & \textbf{17.43}                                  & \textbf{54.96}                                 & 16.22              & \textbf{103.96}                             
 \\
 & RCC~\cite{hobley2022learning}                          & arXiv2022                     & 0      & 17.49                                  & 58.81                                 & 17.12             & 104.53                                 \\
\hline 
%\multicolumn{7}{c}{zero-shot}                                         \\ \hline

\multirow{2}{*}{Zero-shot} & Xu \textit{et al.}~\cite{xu2023zero}                          & CVPR2023                     & 0      & 26.93                                 & 88.63                                 & 22.09              & 115.17                                 \\
& Ours                       & MM2023                             & 0      & \textbf{18.79}                                     & \textbf{61.18}                                      & \textbf{17.78}                                     & \textbf{106.62}                                       \\ \bottomrule
\end{tabular}%
}

\end{table*}

\section{Result and Analysis}
\subsection{Quantitative Result}
To the best of our knowledge, this is the \final{first investigation into} the application of VLMs for zero-shot density map regression in both the generalized counting and class-specific counting settings. Therefore, a direct comparison with previous methods is impractical. To evaluate the performance of our approach, we compare it to several state-of-the-art few-shot and reference-less generalized object counting methods, including FamNet~\cite{ranjan2021learning}, CFOCNet~\cite{yang2021class}, CounTR~\cite{liu2022countr}, LOCA~\cite{djukic2022low}, RepRPN-C~\cite{ranjan2022exemplar}, RCC~\cite{hobley2022learning}, and the only other zero-shot counting method~\cite{xu2023zero} on FSC-147. Additionally, we compare our model with class-specific counting models on the CARPK and ShanghaiTech crowd counting datasets in a cross-dataset setting.

\textbf{Quantitative Result on FSC-147.} We compare CLIP-Count with state-of-the-art generalized counting methods on FSC-147, and summarize quantitative result in Tab.~\ref{tab:quantres}. Overall, we outperform the state-of-the-art zero-shot object counting method~\cite{xu2023zero}
 on the FSC-147 dataset \final{by a significant margin}. We also compare our results with few-shot and reference-less methods for reference. Notice that zero-shot object counting aims to solve a more challenging scenario, as the model needs to understand the \final{correspondence between text} and visual features.

\begin{figure*}
    \centering
    \includegraphics[width=\linewidth]{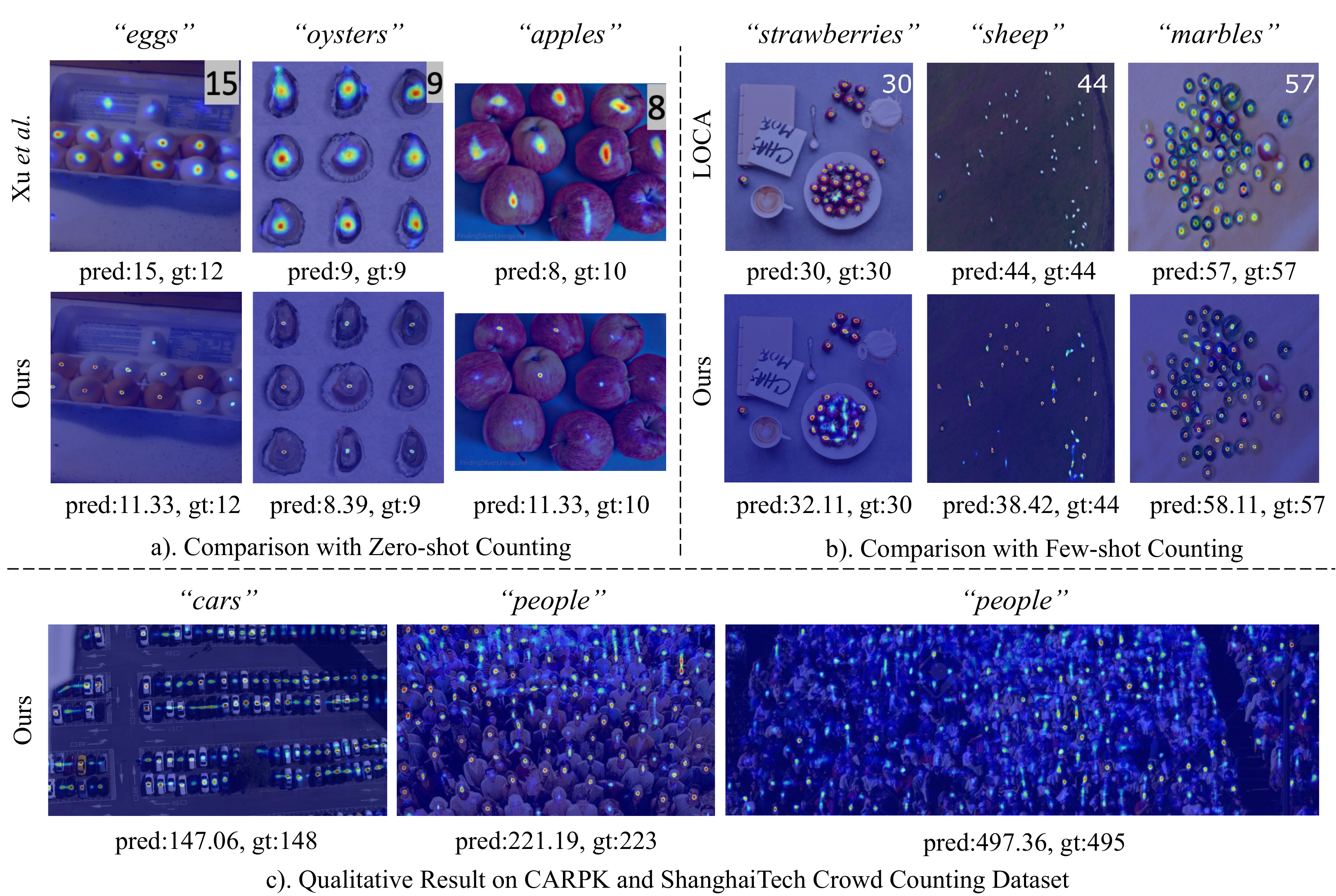}
    \caption[Qualitative Result.]{Qualitative result. We compare CLIP-Count with the state-of-the-art zero-shot ~\cite{xu2023zero} and few-shot counting method~\cite{djukic2022low} on FSC-147 in a) and b), respectively. In c) we visualize results on the CARPK and ShanghaiTech crowd counting datasets.}
    \label{fig:qual_compare}
\end{figure*}

% \begin{table}[t]
% \centering
% \caption{Cross-Dataset Evaluation on CARPK Dataset}
% \label{tab: carpk}
% \begin{tabular}
% {ccccc}
% \hline 
% Method &Finetune& \#Shot 
% &MAE& RMSE  \\ \hline

% FamNet&\checkmark &3 &28.84 & 44.47\\
% CounTR& & 3 & &\\
% RCC& &0& 12.31 & 15.40\\
% CounTR &0 & &\\
% Ours &0 & \textbf{11.96}& 16.61 \\
% \hline
% \end{tabular}%

% \end{table}

\begin{table}[t]
\centering
\caption{Cross-dataset evaluation on CARPK dataset. (\textdagger) Trained using same backbone with ours.}
\label{tab:carpk}
\begin{tabular}
{cccc}
\toprule
Method & \#Shot 
&MAE& RMSE  \\ \hline

FamNet~\cite{ranjan2021learning} &3 &28.84 & 44.47\\
BMNet~\cite{shi2022represent}&  3 &14.41 & 24.60\\
BMNet+~\cite{shi2022represent}&  3 &\textbf{10.44} & \textbf{13.77}\\
\hline
RCC\textsuperscript{\textdagger}~\cite{hobley2022learning} &0& 21.38 & 26.15\\
Ours &0 & \textbf{11.96}& \textbf{16.61} \\
\bottomrule
\end{tabular}%

\end{table}

\begin{table}[t]
\centering
\caption{Cross-dataset evaluation on ShanghaiTech crowd counting dataset. (\textdagger) Trained using same backbone with ours.}
\label{tab:shanghai}

\resizebox{1\columnwidth}{!}{%
\begin{tabular}{ccccccc}
\toprule

Method & Type & Training$\rightarrow$Testing & MAE & RMSE\\
\hline

MCNN~\cite{zhang2016single} & \multirow{2}{*}{Specific} &\multirow{2}{*}{Part A$\rightarrow$Part B} &  85.2&142.3\\
CrowdCLIP~\cite{liang2023crowdclip}  &   &  &69.6 &80.7\\
\cline{2-5}
RCC\textsuperscript{\textdagger}~\cite{hobley2022learning} & \multirow{2}{*}{Generic} & \multirow{2}{*}{FSC147$\rightarrow$Part B}  &66.6&104.8 \\
Ours &   &  &\textbf{45.7}&\textbf{77.4} \\

\hline

MCNN~\cite{zhang2016single}  & \multirow{2}{*}{Specific} &\multirow{2}{*}{Part B$\rightarrow$Part A} &221.4 &357.8 \\
% DConvNet &specific &Part A & 140.4 &357.8 & &\\
CrowdCLIP~\cite{liang2023crowdclip} &   &  &217.0 &322.7\\
\cline{2-5}
RCC\textsuperscript{\textdagger}~\cite{hobley2022learning} & \multirow{2}{*}{Generic} & \multirow{2}{*}{FSC147$\rightarrow$Part A} &240.1 & 366.9  \\
Ours &   &  &\textbf{192.6} &\textbf{308.4} \\
\bottomrule

\end{tabular}%
}
\end{table}

\begin{table*}[]
\caption{Ablate Study. \final{We ablate (A) how to interact multi-modal feature; (B) how to transfer knowledge in CLIP; (C) how to apply contrastive loss.} We show the number of trainable parameters and the evaluation metrics on the FSC-147 dataset.}
\label{tab:ablate}
\centering
\resizebox{1.7\columnwidth}{!}{%
\begin{tabular}{ccccccccc}
\toprule
\multirow{2}{*}{Method} & \multirow{2}{*}{Interaction} & \multirow{2}{*}{Transfer} & \multirow{2}{*}{Contrastive} &\multirow{2}{*}{\#Param.}&\multicolumn{2}{c}{Val Set} & \multicolumn{2}{c}{Test Set} \\
\cline{6-9}
& & & & &  MAE & RMSE & MAE & RMSE

 \\ 

                     \hline
            % \multicolumn{9}{c}{Feature Interaction} \\\hline
A1  & Add & Freeze & None &3.3M& 30.40                           & 96.67                                 & 33.84                  &    124.64              \\

A2 & Naive &Freeze&None&16.0 M&   32.50                &              95.02    & 31.48 & 121.95                    \\

A3      &2-Scale&Freeze&None&   16.6M                         & 29.96                                  & 93.07                                &        28.84            &           120.03         \\\hline
B1&Add&Finetune&None&89.4M&28.79&90.84	&28.87&	120.32\\

B2 &2-Scale&Finetune&None&   102.0M                     & 24.88                                  & 77.73                                 & 22.51                  &    131.12                     \\

B3 &Add	&VPT	&None&4.0M&	22.29 &	67.97  &	20.90	& 110.14 \\

B4                           &Naive&VPT&None&16.6M & 20.43                                       & 69.26                                     & 19.26                                       &      108.71        

\\

B5                           &2-Scale&VPT&None&17.3M &   20.01                                     &    66.08                                   &18.77                                        &     108.00                                                                        \\\hline

C1   &2-Scale&VPT&Visual&                             17.3M& 19.21 &                                       64.93&                                       18.55 &                                       107.51                                        \\
C2                                  &2-Scale&VPT&Multi-modal&17.3M&   \textbf{18.79}                                     & \textbf{61.18}                                      & \textbf{17.78 }                                      & \textbf{106.62}                                       \\
\bottomrule
\end{tabular}%
}

\end{table*}

\begin{figure}
    \centering
    \includegraphics[width=1\linewidth]{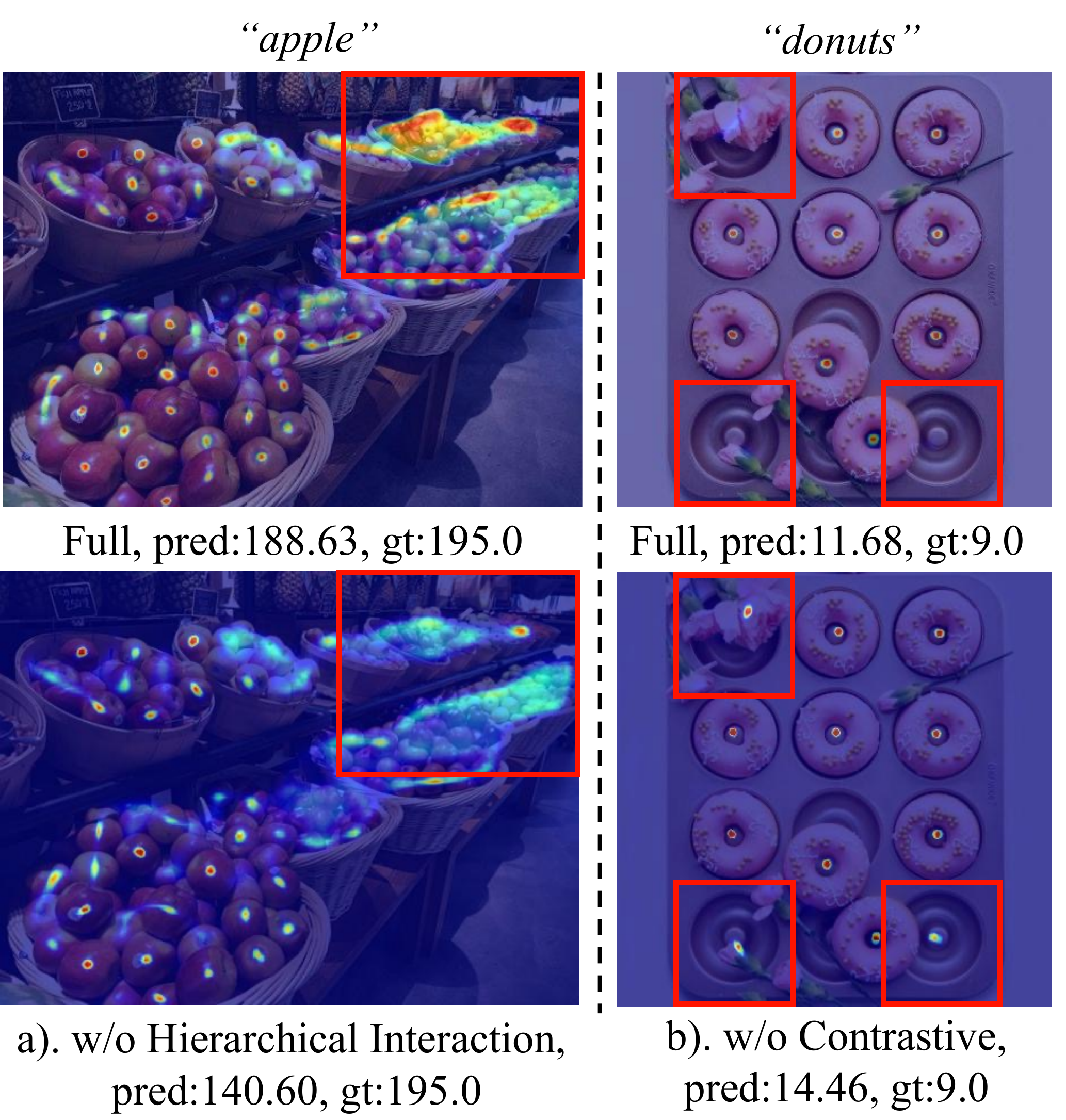}
    \caption{Visualization of ablation study. a:) Replace the hierarchical interaction with a plain transformer; b:) remove
the contrastive loss.}
    \label{fig:ablate}
\end{figure}

\textbf{Quantitative Result on CARPK.} Following previous class-agnostic counting methods ~\cite{lu2018class,hobley2022learning,shi2022represent}, we test CLIP-Count on CARPK to evaluate its cross-dataset generalizability. Specifically, the model is trained on FSC-147 and evaluated on the testing set of CARPK without fine-tuning. To ensure fair comparison, we train RCC with the same visual backbone (i.e.,\ Vit-B/16) on FSC-147, instead of their proposed FSC-133.  As shown in Tab.~\ref{tab:carpk}, our approach outperforms the representative reference-less counting method, RCC, by a significant margin.

\textbf{Quantitative Result on ShanghaiTech \final{Crowd Counting}.} Concurrent with this work, Liang \textit{et al.}\ ~\cite{liang2023crowdclip} propose the first CLIP-based crowd counting method, namely CrowdCLIP. In their paper, they evaluate the cross-dataset performance of models by training the class-specific models on one part of the ShanghaiTech dataset and test on another. We conduct a similar experiment by directly evaluating our model and RCC~\cite{hobley2022learning} for crowd counting (\textit{i.e.},\ no finetuning on any part of ShanghaiTech). Similar as the experiment on CARPK, we have also modified the backbone of RCC to ensure fairness. The results are summarized in Tab.~\ref{tab:shanghai}. Our proposed generic counting method outperforms the representative class-specific counting methods MCNN~\cite{zhang2016single} and CrowdCLIP~\cite{liang2023crowdclip} by a significant margin, even without training on any part of the crowd-counting dataset.  In particular, we outperform CrowdCLIP by 11.2\%, 4.0\%, 34.3\%, and 3.7\% under cross-dataset setting. Moreover, our method significantly outperforms the other reference-less counting method, RCC~\cite{hobley2022learning}. Overall, the experiment demonstrate the advanced cross-dataset generalizability of CLIP-Count.
 
\subsection{Qualitative Result}

We visualize qualitative results in Fig.~\ref{fig:qual_compare}, where we overlay the predicted density map on the input image. We provide a side-by-side comparison with our method against the state-of-the-art zero-shot counting method~\cite{xu2023zero} in Fig.~\ref{fig:qual_compare}-(a), and few-shot method LOCA~\cite{djukic2022low}, in Fig.~\ref{fig:qual_compare}-(b) on the FSC-147 dataset from their paper. Compared with~\cite{xu2023zero},  our proposed method could better localize high density in the center of the object with high-fidelity.  By contrast, the density prediction in~\cite{xu2023zero} \final{may} exhibit non-concentrated patterns. Our model could also achieve competitive results with the representative few-shot counting method~\cite{djukic2022low}, both in terms of prediction error and fidelity of density maps. Additionally, we visualize \final{results of crowded scenes} from the CARPK and ShanghaiTech datasets \final{in Fig.~\ref{fig:qual_compare}-(c).} \final{Overall, the proposed method robustly localizes and counts objects of different categories, shapes, sizes, and densities.}

\subsection{Ablation Study}
We validate the design choices \final{by iteratively adding components with different designs to our baseline model (A1)}. Specifically, we have ablated with the following three design choices: (A) how to interact \final{text and image features from CLIP}; (B) how to fine-tune CLIP ViT; and (C) the effect of patch-text contrasitive loss. For (A), \textit{``add''} means add text embedding to patch embeddings, \textit{``naive''} means use a plain vision transformer~\cite{dosovitskiy2020image,liu2022countr} without hierarchical design; for (B) \textit{``finetune''} means finetune all ViT parameters; for (C) Contrastive loss, \textit{``visual''} means adjust patch embeddings only, and \textit{``multi-modal''} means adjust text and patch embeddings at the same time. We summarize the results in Tab.~\ref{tab:ablate}, and we also visualize the effect of hierarchical interaction and contrastive loss in Fig.~\ref{fig:ablate}.

The proposed designs effectively improve the performance of \final{our model}. The cross-attention  outperforms baseline \final{(A1)} in modeling complex text-image relationships.  The hierarchical design of \final{the} interaction module \final{further} helps the model handle size variations of objects, such as the example in Fig.~\ref{fig:ablate}-(a). Additionally, the VPT enables parameter-and-data-efficient transfer learning, bringing 19.5\% and 14.5\% gains in MAE and RMSE \final{(A3 \textit{vs}. B5)}, respectively, with only 3\% of additional parameters.  The contrastive loss guides the model to \final{align} text and dense image features, resulting in improvements of 6.1\% and 7.6\% for both metrics \final{(B5 \textit{vs}. C2)}.  Without the contrastive loss, the model may mistakenly treat visually similar objects as counting target, as \final{exemplified} in Fig.~\ref{fig:ablate}-(b).

\subsection{Limitations and Future Works} \label{sec: fail}

% Despite its promising results, CLIP-Count may encounter limitations in certain scenarios due to the inherent ambiguity in text guidance for object counting. Two representative examples are visualized in Fig.~\ref{fig: fail}. Specifically, in (a), the prompt \textit{``apple''} can refer to either the fruit or a drawing of an apple, presenting a case of semantic ambiguity. \final{When encounter} linguistic ambiguity like this, our model may produce unwanted results. Another example is shown in Fig.~\ref{fig: fail}-(b), where the queried object itself could be ambiguous, as the \textit{``sunglasses''} consists of two similar components (i.e., two lenses). While the ground truth treats the whole pair as one object, our model counts them separately, resulting in a doubled count. We attribute these limitations to the inadequate prompt annotation in FSC-147, where only a general class name is available for text guidance. In future work, we plan to focus on collecting a dataset with more fine-grained text annotation to disambiguate the query object and enhance model accuracy.

\begin{figure}
    \centering
    \includegraphics[width=\linewidth]{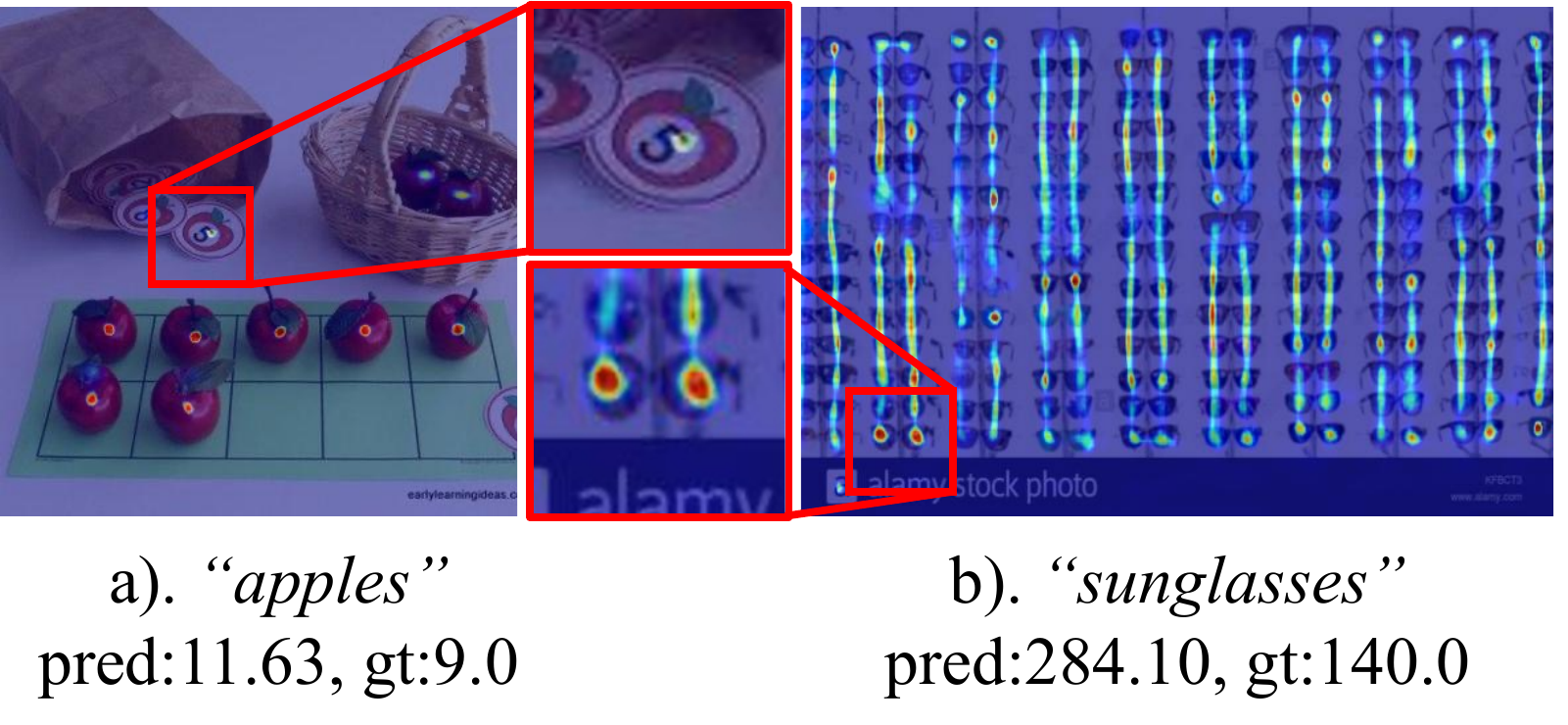}
    \caption[Failure Cases of CLIP-Count.]{\textbf{Representative failure cases due to \final{a:) semantic ambiguity of the input prompt, b:). self-repeating structure of the queried object.}}}
    \label{fig: fail}
\end{figure}

\final{CLIP-Count can be limited in certain scenarios due to ambiguity in text guidance for object counting. For example in Fig.~\ref{fig: fail}-(a)}, the prompt \textit{``apple''} can refer to either the fruit or a drawing of an apple, presenting a case of semantic ambiguity. \final{When encounter} linguistic ambiguity like this, our model may produce unwanted results. Another example is shown in Fig.~\ref{fig: fail}-(b), where the queried object itself could be ambiguous, as the \textit{``sunglasses''} consists of two similar \final{lens, making it unclear whether they should be counted together or separately}. We attribute these limitations to the inadequate prompt annotation in FSC-147, where only a general class name is available for text guidance. In future work, we plan to focus on collecting a dataset with more fine-grained text annotation to disambiguate the query object and enhance model accuracy.

\section{Conclusion}
% This paper presents CLIP-Count, a novel generalized object counting method that leverages text guidance to achieve zero-shot generalized object counting. The proposed method leverages the recently proposed visual-language model, CLIP, to enable zero-shot image-text matching.  To transfer image-level CLIP for dense tasks, a patch-text contrastive loss is introduced to align text with dense visual features, enabling parameter-and-data-efficient transfer of pre-trained knowledge in CLIP to the task of dense prediction through visual prompt tuning.  To fuse semantic information to the image feature at scale, a hierarchical patch-text interaction module is introduced, followed by a CNN-based decoder.  Experimental results on the FSC-147 dataset demonstrate that our method outperforms the current state-of-the-art zero-shot generalized counting method. Further experiments on the CARPK and ShanghaiTech crowd counting datasets further demonstrate the generalizability of proposed method in a cross-dataset setting. We have also experimented with the key design choices in CLIP-Count and demonstrated the effectiveness of our design. However, failure cases are observed, particularly due to inherent ambiguity in text, and we suggest future research focus on establishing a counting dataset with fine-grained text annotations.

\final{In this paper, we show that pre-trained VLMs could be directly adopted for text-guided object counting tasks in an end-to-end manner. Specifically, we propose CLIP-Count, which fully exploits the pretrained knowledge in CLIP by aligning text embedding with patch-level visual features. Extensive experiments on the FSC-147, CARPK and ShanghaiTech crowd counting datasets demonstrate the state-of-the-art accuracy and generalizability of the proposed method for zero-shot object counting.}

%\begin{acks}
%To Robert, for the bagels and explaining CMYK and color spaces.To Robert, for the bagels and explaining CMYK and color spaces.To Robert, for the bagels and explaining CMYK and color spaces.To Robert, for the bagels and explaining CMYK and color spaces.
%\end{acks}

\bibliographystyle{ACM-Reference-Format}
\balance
\bibliography{main}

\clearpage
\appendix

\section{Further Ablations}
We ablate the number of resolution scales in our proposed hierarchical patch-text interaction module. The result is summarized in Tab.~\ref{tab:append_ablate}. Specifically, for the 1-Scale interaction (\textit{i.e.}, B4 in Tab.~\ref{tab:ablate}), we use a plain ViT~\cite{dosovitskiy2020image} with 4 layers and 4 head to maintain similar trainable parameters with our 2-Scale interaction (\textit{i.e.}, B5 in Tab~\ref{tab:ablate}). For the 3-Scale interaction, the finest spatial resolution of the feature map is 
$4p \times 4p$, which is added to the 3rd layer of the density decoder in a similar way as Eqn.~\ref{eq: hierachy}. We observe that the performance gradually gets better as the number of interactions increases. Using 2-Scale interaction, our method achieves highly competitive results, while more interactions with more parameters no longer lead to significant performance improvement.

\begin{table}[h]
\centering
\caption{Ablation on number of scales in hierarchical patch-text interaction module on FSC-147.}
\label{tab:append_ablate}

\resizebox{1\columnwidth}{!}{%
\begin{tabular}{cccccc}
\toprule
Interaction&\#Param.&Val MAE&Val RMSE&Test MAE&Test RMSE
\\
\hline

1-Scale&16.6M&20.43&69.26&19.26&108.71\\
2-Scale&17.3M&20.01&66.08&18.77&108.00\\
3-Scale&24.0M&
19.81&66.36&19.00&104.44\\
\bottomrule

\end{tabular}%
}
\end{table}

\section{Inference Details}

Similar with prior work~\cite{liu2022countr}, we apply sliding window during inference time to deal with variable image input size. To be more specific, given an image $I \in \mathbb{R}_{H\times W\times3}$ we resize it $I^\prime \in \mathbb{R}_{H^\prime\times W^\prime \times3}$, to make the shortest side becoming 224 pixels while maintaining aspect ratio: 
\begin{equation}
    \operatorname{min}(H^\prime,W^\prime) = 224, HW^\prime=H^\prime W
\end{equation}
Then we apply sliding window algorithm on $I^\prime$ with window size = $224\times 224$ and stride $ = 128$. The predicted density estimation at the overlapped area are averaged.

\section{Additional Qualitative Results}
\begin{figure*}
    \centering
    \includegraphics[width=0.85\linewidth]{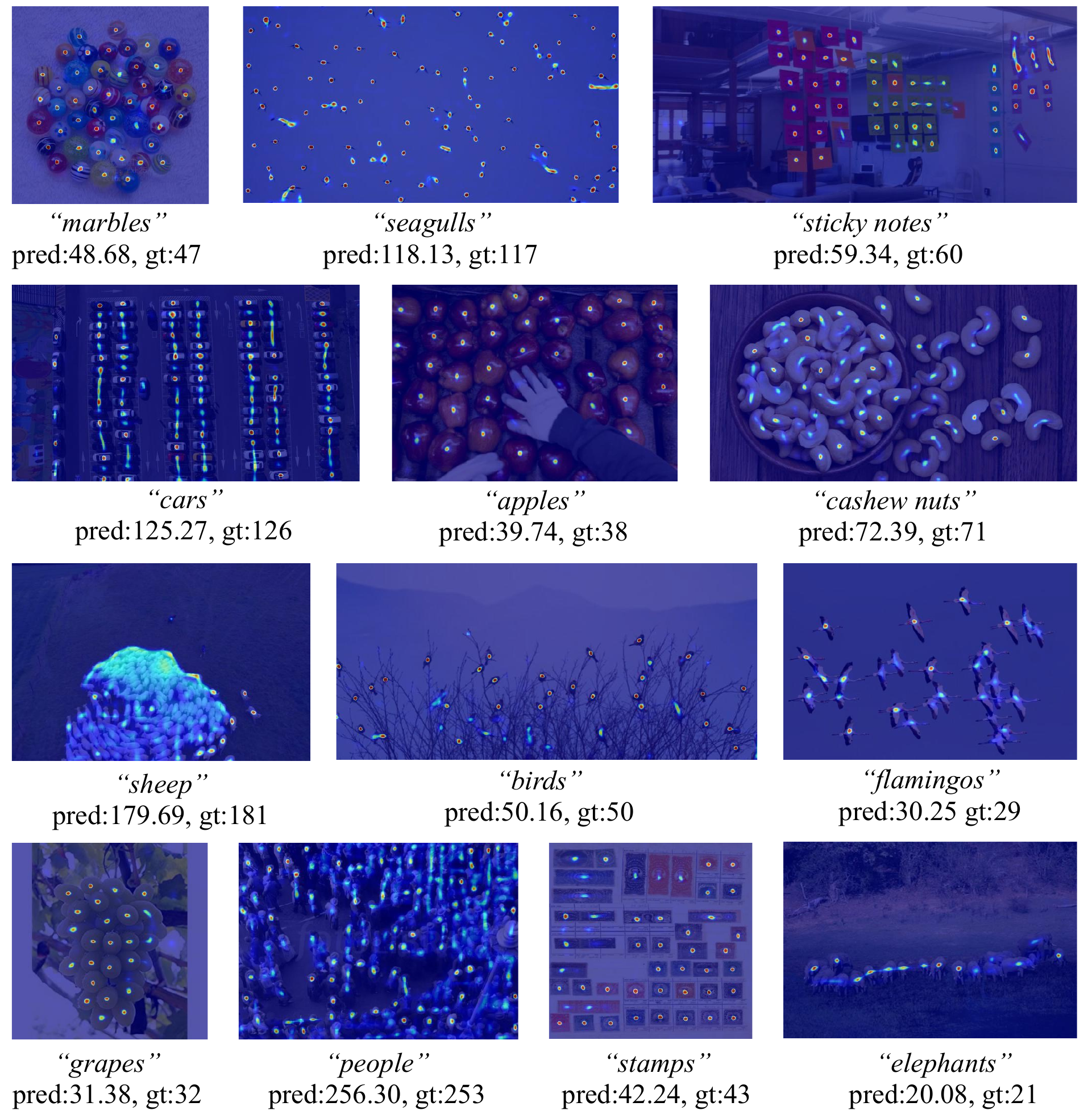}
    \caption{Additional qualitative results. CLIP-Count could generalize well to object with different categories, shapes, sizes, and densities. }
    \label{fig:more_qual}
\end{figure*}

We show more qualitative results on FSC-147, CARPK, and ShanghaiTech Crowd Counting dataset in Fig.~\ref{fig:more_qual}.

\end{document}